\documentclass[10pt,twocolumn,letterpaper]{article}

\usepackage[pagenumbers]{cvpr} 
\usepackage[table]{xcolor} 
\usepackage{colortbl}      
\usepackage{graphicx}
\usepackage{capt-of} 
\usepackage{float}
\usepackage{pifont}
\usepackage{multirow}
\definecolor{cvprblue}{rgb}{0.21,0.49,0.74}
\usepackage[pagebackref,breaklinks,colorlinks,allcolors=cvprblue]{hyperref}

\definecolor{redcell}{HTML}{AE282C}
\definecolor{greencell}{HTML}{D47264}
\definecolor{yellowcell}{HTML}{F6D6C2}

\newcommand{\redbg}{\cellcolor{redcell!40}}     
\newcommand{\greenbg}{\cellcolor{greencell!30}}
\newcommand{\yellowbg}{\cellcolor{yellowcell!30}}
\def\paperID{42696} 
\def\confName{CVPR}
\def\confYear{2026}

\title{NeVStereo: A NeRF-Driven NVS-Stereo Architecture for High-Fidelity 3D Tasks}

\author{Pengchen Chen\footnotemark[1]\hspace{0.4em}\footnotemark[2]\\
University of Washington\\
{\tt\small pengcc@uw.edu}
\and
Yue Hu\footnotemark[1]\hspace{0.4em}\footnotemark[2]\\
University of Southern California\\
{\tt\small yhu57782@usc.edu}
\and
Wenhao Li\footnotemark[1]\hspace{0.4em}\footnotemark[3]\\
CUHK-Shenzhen\\
{\tt\small wenhaoli1@link.cuhk.edu.cn}
\and
Nicole M Gunderson\\
University of Washington\\
{\tt\small nmgundo@uw.edu}
\and
Andrew Feng\\
USC Institute for Creative Technologies\\
{\tt\small feng@ict.usc.edu}
\and
 Zhenglong Sun\\
CUHK-Shenzhen\\
{\tt\small sunzhenglong@cuhk.edu.cn}
\and
Peter Beerel\\
University of Southern California\\
{\tt\small pabeerel@usc.edu}
\and
Eric J Seibel\\
University of Washington\\
{\tt\small eseibel@uw.edu}
}

\begin{document}

\twocolumn[{%
    \renewcommand\twocolumn[1][]{#1}%
    \maketitle
    \begin{center}
        \vspace{-3mm} 
        \includegraphics[width=1\linewidth]{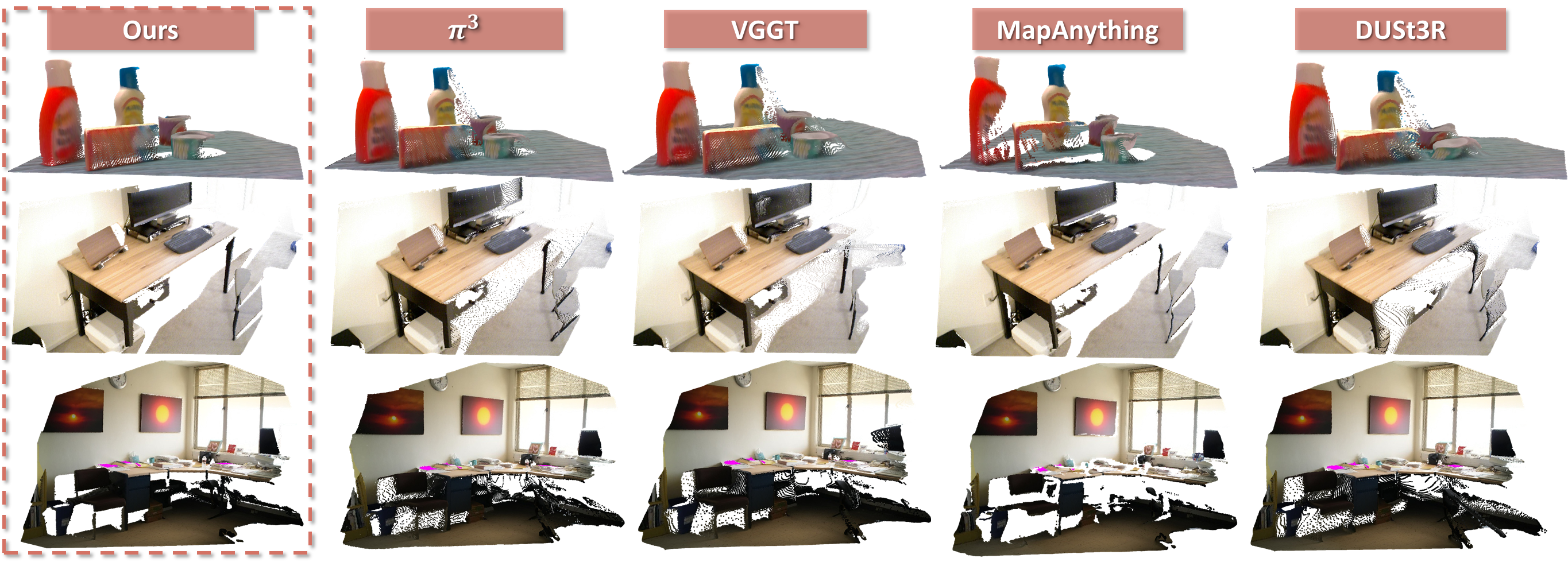}
        \captionof{figure}{3D reconstruction quality comparison. NeVStereo produces point clouds with more accurate geometry and significantly fewer floating artifacts than other methods.}
        \label{fig:pointcloud}
        \vspace{3mm} 
    \end{center}
}]

\renewcommand{\thefootnote}{\fnsymbol{footnote}}
\footnotetext[1]{Co-first authors, equal contribution.}
\footnotetext[2]{Primary contributors to the main methods and evaluation.}
\footnotetext[3]{Contributed to preliminary study.}
\renewcommand{\thefootnote}{\arabic{footnote}}

\begin{abstract}


In modern dense 3D reconstruction, feed-forward systems (e.g., VGGT, $\pi^3$) focus on end-to-end matching and geometry prediction but do not explicitly output the novel view synthesis (NVS). Neural rendering-based approaches offer high-fidelity NVS and detailed geometry from posed images, yet they typically assume fixed camera poses and can be sensitive to pose errors. As a result, it remains non-trivial to obtain a single framework that can offer accurate poses, reliable depth, high-quality rendering, and accurate 3D surfaces from casually captured views. \textbf{We present NeVStereo}, a NeRF-driven NVS-stereo architecture that aims to jointly deliver camera poses, multi-view depth, novel view synthesis, and surface reconstruction from multi-view RGB-only inputs. NeVStereo combines NeRF-based NVS for stereo-friendly renderings, confidence-guided multi-view depth estimation, NeRF-coupled bundle adjustment for pose refinement, and an iterative refinement stage that updates both depth and the radiance field to improve geometric consistency. This design mitigated the common NeRF-based issues such as surface stacking, artifacts, and pose-depth coupling. Across indoor, outdoor, tabletop, and aerial benchmarks, our experiments indicate that NeVStereo achieves consistently strong zero-shot performance, with up to \textbf{36\% lower depth error}, \textbf{10.4\% improved pose accuracy}, \textbf{4.5\% higher NVS fidelity}, and state-of-the-art mesh quality (\textbf{F1 91.93\%, Chamfer 4.35 mm}) compared to existing prestigious methods.
\end{abstract}

\section{Introduction}
\label{sec:intro}

Modern 3D vision increasingly demands reconstruction systems that deliver accurate geometry, reliable camera poses, and photorealistic novel view synthesis (NVS) from casually captured multi-view images~\cite{kerbl20233d,mildenhall2021nerf,wang2024dust3r,cabon2025must3r}, for applications such as robotics, AR/VR~\cite{tonderski2024neurad}, cultural heritage digitization~\cite{warburg2023nerfbusters,croce2024comparative}, and image-guided surgery~\cite{chen2024endoperfect,zhou2019real}. This raises a central question: can we design a reconstruction paradigm that simultaneously provides high-accuracy 3D geometry, robust pose estimation, and high-quality NVS across diverse scene scales, baselines, and capture conditions?

Recent progress toward this goal has largely followed two separate lines. Feed-forward 3D systems such as MaSt3R~\cite{leroy2024grounding}, VGGT~\cite{wang2025vggt}, and $\pi^{3}$~\cite{wang2025pi} focus on end-to-end matching, geometry reconstruction, and pose estimation from dense multi-view images, offering strong geometric priors but no explicit NVS module. Neural rendering–based methods (e.g., 2DGS~\cite{2dgs}, Neuralangelo~\cite{li2023neuralangelo}, NeuS~\cite{wang2021neus}, SUGAR~\cite{guedon2024sugar}) instead optimize a neural scene representation from posed images for high-quality NVS and 3D reconstruction, yet typically keep camera poses fixed and remain sensitive to pose errors. 


To bridge the missing components of these two lines, we consider an alternative paradigm, \emph{NVS-stereo}: use NVS to render high-fidelity stereo pairs, then apply a powerful stereo depth estimator to the synthesized pairs, and finally feed the resulting depth back to refine poses, NVS quality, and a fused 3D mesh. This idea is not purely conceptual: prior work has already combined 3D Gaussian Splatting (3DGS) with NVS-stereo for depth estimation and reconstruction~\cite{wolf2024gs2mesh,safadoust2024self}, but with poses fixed to the original SfM solution. To understand whether this limitation stems from NVS-stereo itself or from the underlying representation, we conduct a systematic comparison between 3DGS- and NeRF-based NVS-stereo pipelines. Our analysis shows that 3DGS-based NVS-stereo quickly saturates at the SfM accuracy ceiling, whereas NeRF-based NVS-stereo can surpass this ceiling by producing smoother, more accurate, and more multi-view-consistent depth.

Building on these insights, we propose \textbf{NeVstereo}: a complete NeRF-based NVS-Stereo framework that integrates novel view synthesis, stereo depth estimation, pose optimization, depth fusion, and iterative refinement (Fig. ~\ref{fig:framework}). It effectively addresses all the aforementioned challenges. Our method achieves state-of-the-art performance in camera pose estimation, multi-view depth estimation, novel view synthesis, and surface reconstruction, and surpasses most prestigious existing methods. Specifically, our \textbf{contributions} are as follows:

\begin{itemize}




    \item \textbf{NeRF-based high-accuracy NVS-Stereo reconstruction.} We introduce the first NeRF-based NVS-stereo architecture that achieves high-accuracy pose, depth novel view synthesis, and 3D reconstruction at the same time, through our designed pose optimization and NeRF iterative pipeline. By using refined depths and poses to focus sampling on true surfaces and filtering NeRF artifacts with confidence-aware stereo cues, our method yields more geometrically consistent NVS, supports larger baselines, and delivers highly accurate depth and 3D geometry.

    \item \textbf{Systematic study of NVS-Stereo reconstruction.} We provide the statistical analysis of NVS-Stereo pipelines, revealing core failure modes and, through controlled 3DGS-vs.-NeRF comparisons, demonstrating that 3DGS-based NVS-Stereo saturates at the SfM ceiling and when NeRF can surpass it.

    \item \textbf{Strong empirical performance and generalization.} Across indoor/outdoor, tabletop and UAV benchmarks, our architecture outperforms SOTA baselines, achieving up to 36\% lower \textbf{depth} error, 10.4\% better \textbf{pose} accuracy (RTE 0.0215 vs. 0.0240), 4.5\% higher \textbf{NVS render} PSNR, and best-in-class \textbf{mesh} quality (F1 91.93\% vs. 89.30\%; Chamfer 4.35 mm vs. 4.74 mm).

\end{itemize}

\section{Related Works}
\label{sec:related}

\textbf{Novel View Synthesis:}
Synthesizing high-quality stereo view pairs from known viewpoints is the cornerstone of our pipeline. Among NeRF-based NVS methods, ZipNeRF~\cite{barron2023zip} remains state-of-the-art across diverse datasets, producing anti-aliased, view-consistent renderings with stable geometry that are well suited for downstream stereo. It combines anti-aliased volumetric rendering, multi-scale supervision, and proposal-driven sampling with a coarse-to-fine training schedule, yielding sharper textures, fewer floaters, and improved handling of thin and specular structures.

\textbf{Stereo depth estimation:}
Stereo has evolved from attention/transformer and deformable-CNN pipelines (e.g., MatchStereo~\cite{jing2024match}, DEFOM-Stereo~\cite{jiang2025defom}) to large, pre-trained foundation models such as FoundationStereo~\cite{wen2025foundationstereo}. While MatchStereo and DEFOM-Stereo offer strong correspondence filtering within their training domains, they generally require task- or dataset-specific tuning under photometric shifts or weak texture. In contrast, FoundationStereo, trained at scale with monocular priors, achieves strong zero-shot generalization and remains robust under low texture, specularities, illumination changes, and the syn-to-real gap introduced by NVS rendering.

\textbf{RGB-D optimization:}
Classical geometric RGB-D SLAM systems such as ORB-SLAM~\cite{mur2015orb,campos2021orb}, BAD-SLAM~\cite{schops2019bad}, and DROID-SLAM~\cite{teed2021droid} jointly estimate poses and structure from photometric and geometric residuals via robust data association and bundle adjustment. Neural rendering–based SLAM instead optimizes camera trajectories against a learned scene representation: GS-SLAM~\cite{yan2024gs}, SplaTAM~\cite{keetha2024splatam} and SplatMAP~\cite{hu2025splatmap} use 3D Gaussian Splatting for efficient mapping, while NeRF-SLAM~\cite{rosinol2023nerf} maintains a volumetric radiance field that provides dense photometric gradients. In this work, we combine these two paradigms by optimizing camera poses under geometric constraints while simultaneously refining rendering quality through neural rendering supervision (Sec.~\ref{sec:method}).

\section{Preliminary Study and Challenges}
\label{sec:preliminary}

To compare the NeRF backbone and the 3DGS backbone for NVS-stereo, we evaluate both on the NVIDIA-HOPE\cite{tyree2022hope} and Redwood RGB-D\cite{Park2017} datasets. We first conduct a cross-dataset analysis using Redwood RGB-D and NVIDIA HOPE to examine how differences in SfM quality affect overall NVS-stereo performance. We then perform an in-depth study within NVIDIA HOPE to analyze how the precision of SfM initialization influences NeRF and 3DGS-based NVS-stereo results. 

We use Nerfacto (without pose optimization) and vanilla 3DGS as backbones, both initialized from COLMAP. Depth maps are obtained by rendering training views as left images, synthesizing novel right views with a fixed baseline, and running FoundationStereo for stereo estimation. Hyperparameters are tuned on 20\% of scenes and fixed for the remaining 80\%. Training stops when PSNR $\geq 35$. As shown in Tab.~\ref{tab:preliminary_table}, we found that the NeRF-based NVS-Stereo performance outperforms the COLMAP and 3DGS-based methods on both absrel and $\delta<5\%$, meanwhile, the 3DGS-based NVS-Stereo didn't surpass COLMAP's performance.

\begin{table}[h]
\centering
\scriptsize
\setlength{\tabcolsep}{4pt}
\renewcommand{\arraystretch}{1.05}
\begin{tabular}{c|cc|cc}
\hline
\textbf{Method} & \multicolumn{2}{c|}{\textbf{NVIDIA-HOPE}} & \multicolumn{2}{c}{\textbf{Redwood RGB-D}} \\
\cline{2-5}
 & absrel$\downarrow$ & $\delta<5\%\uparrow$ & absrel$\downarrow$ ($\Delta$\%) & $\delta<5\%\uparrow$ ($\Delta$\%) \\
\hline
COLMAP   & 0.019 & 93.80\% & 0.048 {\scriptsize(+152.4\%)} & 79.94\% {\scriptsize(-14.8\%)} \\
3DGS based & 0.023 & 94.32\% & 0.048 {\scriptsize(+110.9\%)} & 82.21\% {\scriptsize(-12.8\%)} \\
Nerfacto based & 0.018 & 95.47\% & 0.023 {\scriptsize(+33.7\%)} & 91.08\% {\scriptsize(-4.6\%)} \\
\hline
\end{tabular}
\caption{Depth accuracy on NVIDIA-HOPE and Redwood RGB-D. NeRF-based NVS-Stereo performs best on both metrics, while 3DGS-based NVS-Stereo does not surpass COLMAP and degrades more under weakened SfM initialization.}
\label{tab:preliminary_table}
\end{table}



\begin{figure}
    \centering
    \includegraphics[width=1\linewidth]{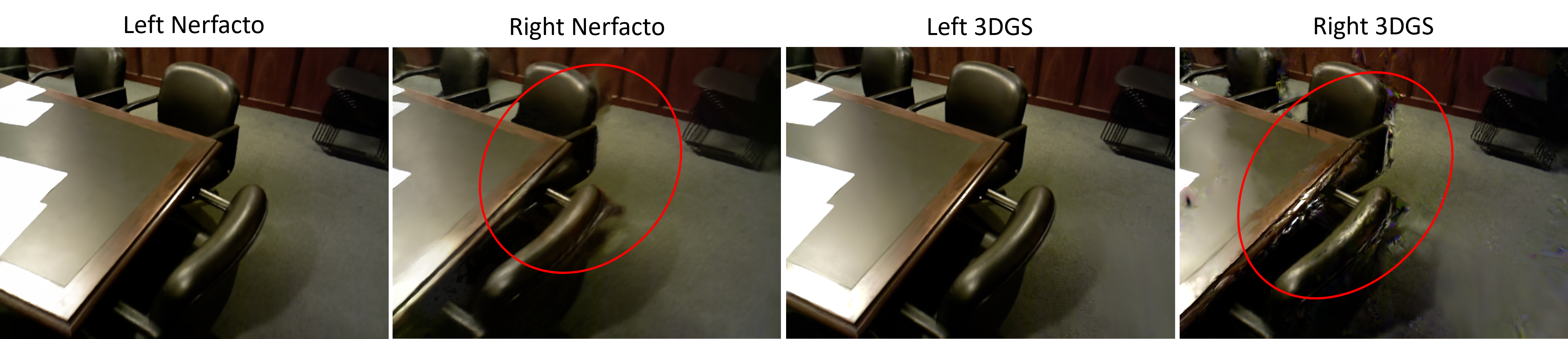}
    \caption{Artifact comparison under degraded SfM initialization. While input-view renderings look similar, the novel views show that 3DGS produces sharper, structured artifacts than NeRF. These artifacts disrupt stereo correspondence and explain the larger accuracy drop of 3DGS-based NVS-stereo.}
    \label{fig:GSvsNeRF}
\end{figure}

When SfM quality degrades across datasets (Tab.~\ref{tab:preliminary_table}), 3DGS accuracy drops nearly as much as COLMAP, while NeRF remains more stable. To investigate this difference, we examine frames with the largest performance gap in Fig.~\ref{fig:GSvsNeRF}. While the left (training) views appear nearly identical, the right (novel) views reveal that 3DGS produces sharper, structured artifacts that disrupt stereo correspondence, whereas NeRF's low-frequency artifacts are less harmful to matching.

\begin{figure*}[!t]
    \includegraphics[width=1\linewidth]{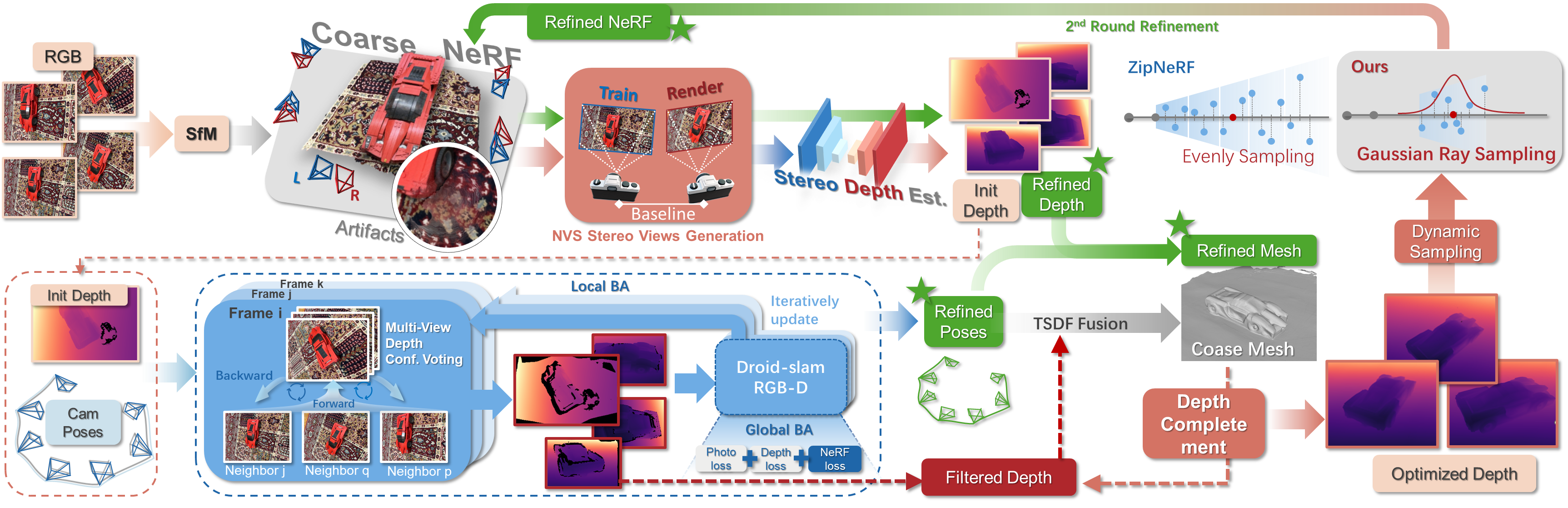}
    \caption{Architecture overview. Multi-view RGB inputs yield an initial SfM reconstruction and a coarse NeRF, which renders stereo pairs for depth estimation. The resulting depths are refined using our modified DROID-SLAM with multi-view depth voting and a NeRF-guided reprojection loss, followed by TSDF fusion and depth completion. The optimized depths then supervise a second-round NeRF refinement using our depth-guided Gaussian ray sampling for improved geometric accuracy. \textcolor[HTML]{0eb83a}{\ding{72} Green highlights denote our outputs.}
}
    \label{fig:framework}
\end{figure*}

On NVIDIA-HOPE, we relate pose error to the fraction of pixels with $\delta<5\%$, shown in Tab.~\ref{tab:pose-corr}. Within $\pm1\sigma$ of the pose-error distribution, neither NeRF-based nor 3DGS-based NVS-Stereo shows a significant correlation, meaning that the depth gap is not pose-driven in the typical regime, in this portion, both NeRF and 3DGS performs a good rendering quality, depth estimation algorithm dominants the performance. However, beyond $1\sigma$, correlations increase for both. NeRF-based NVS-Stereo shows only weak dependence (Pearson $r\approx-0.24$ across ATE/RRE/RTE), whereas 3DGS-based shows moderate dependence ($r\approx-0.46$ in ATE and $r\approx-0.3$ in RRE), consistently stronger than NeRF-based. More analysis diagrams are shown in \textbf{the supplementary material.}

\begin{table}[t]
\centering
\scriptsize
\setlength{\tabcolsep}{4pt}
\renewcommand{\arraystretch}{0.95}
\begin{tabular}{c|cc|cc|cc}
\hline
\textbf{Method} & \multicolumn{2}{c|}{\textbf{ATE}} & \multicolumn{2}{c|}{\textbf{RTE}} & \multicolumn{2}{c}{\textbf{RRE}} \\
\cline{2-7}
 & Pearson & Spearman & Pearson & Spearman & Pearson & Spearman \\
\hline
3DGS$^{\mathrm{N}}$     & -0.071 & -0.050 & -0.011 & -0.038 & -0.025 & -0.043 \\
Nerfacto$^{\mathrm{N}}$ & -0.059 & -0.049 & -0.026 & -0.034 & -0.044 & -0.062 \\
\hline
3DGS$^{\mathrm{O}}$     & \textbf{-0.463} & \textbf{-0.461} & -0.170 & -0.175 & -0.283 & -0.294 \\
Nerfacto$^{\mathrm{O}}$ & -0.238 & -0.267 & -0.215 & -0.143 & -0.169 & -0.089 \\
\hline
\end{tabular}
\caption{Correlation between depth accuracy ($\delta<5\%$ fraction) and pose error on NVIDIA-HOPE. Superscripts mark pose-error regimes: $^{\mathrm{N}}$ within $1\sigma$ (normal), $^{\mathrm{O}}$ greater than $1\sigma$ (outlier). RTE = RPE$_\text{trans}$ ($\delta\!=\!1$); RRE = RPE$_\text{rot}$ ($\delta\!=\!1$).}
\label{tab:pose-corr}
\end{table}

Based on these experiments, we offer the following explanation: 3DGS, as a semi-dense representation initialized from SfM points, inherits both the strengths and weaknesses of the underlying reconstruction. It effectively behaves like a dense extension of COLMAP’s PatchMatch stereo, with performance tightly coupled to SfM quality. In under-constrained regions, its semi-dense nature tends to produce high-frequency, blocky artifacts with sharp discontinuities, which corrupt the cost volume and severely disrupt stereo correspondence. By contrast, NeRF learns a continuous, dense field with smoothly varying texture and geometry across neighboring rays, yielding a cleaner cost volume with sharper dominant peaks and more stable stereo matching. NeRF depends only on camera poses rather than SfM point clouds, and is therefore less rigidly tied to SfM’s global geometry. In regions with pose bias, it tends to favor local photometric consistency over strict global accuracy, preserving high local fidelity at the cost of slight global misalignment. Its artifacts are predominantly low-frequency, fog-like structures instead of sharp discontinuities, which are generally less harmful to modern stereo estimators and thus more favorable for depth estimation.

\begin{figure}[H]
    \centering
    \includegraphics[width=\linewidth]{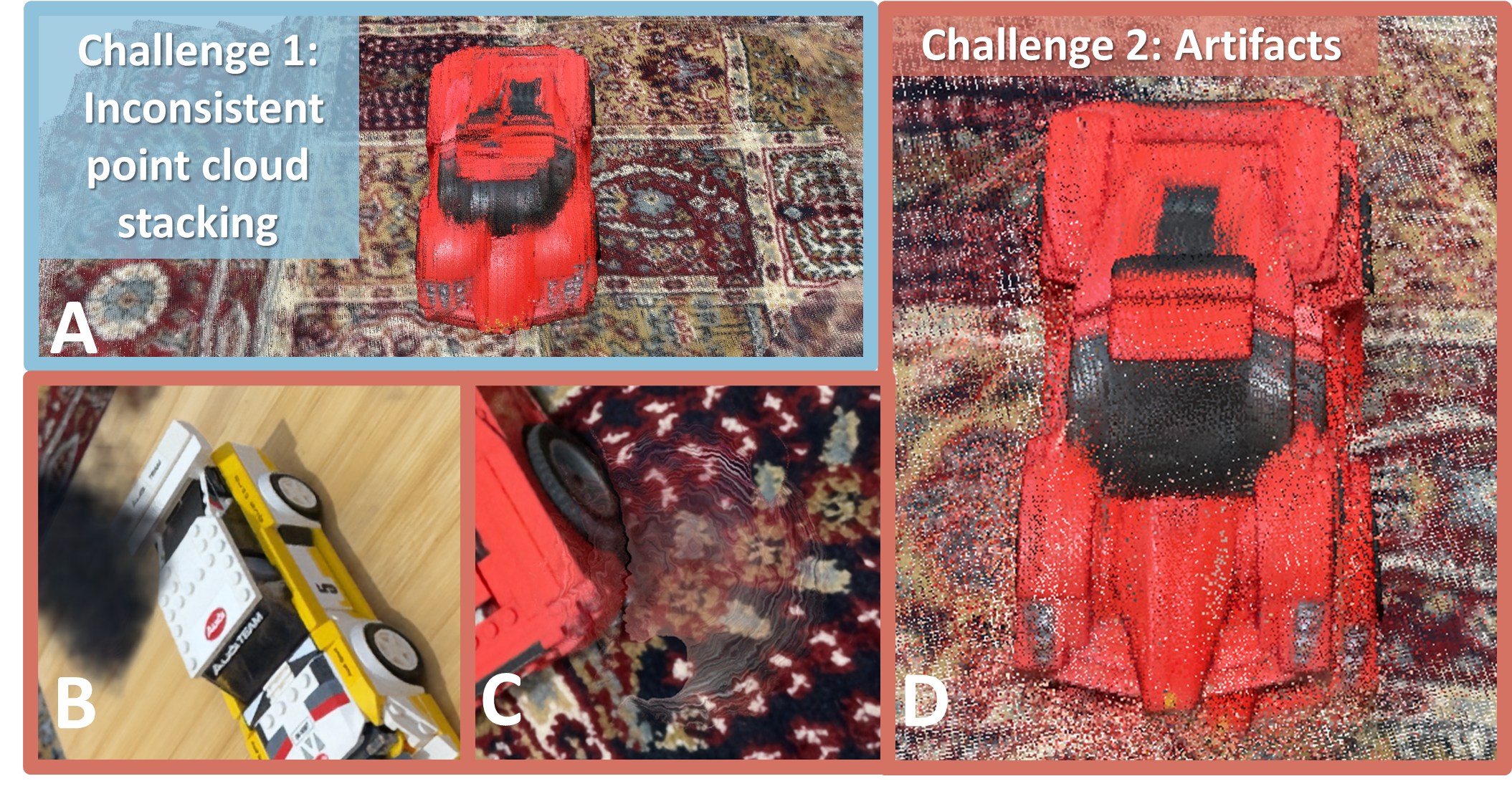}
    \caption{Stereo-depth projection without explicit multi-view constraints produces non-coincident, layered surfaces stacking. Even with accurate per-pixel depths, NeRF’s geometry can drift/ghost across views, breaking cross-view consistency. This \textbf{cannot be solved} by pose-optimized NeRF (BARF or CamP).}
    \label{fig:challenges}
\end{figure}

Therefore, within an NVS-stereo pipeline, NeRF can provide better single-frame depth accuracy than 3DGS. However, this does not make NeRF an overall better backbone for 3D tasks: \textbf{it introduces several stringent challenges due to its lack of an explicit multi-view consistency constraint} that constrains the practical deployment of a NeRF-based NVS-Stereo pipeline:

\textbf{(1) Inconsistent point cloud stacking:}
Point-cloud stacks from NeRF-based NVS-stereo can exhibit noticeable inconsistencies (Fig.~\ref{fig:challenges}A), This due to the NeRF relies only on SfM poses without explicit geometric regularization, so NVS-stereo depths from different viewpoints may not coincide on the same surface points. Pose-refinement methods such as BARF~\cite{lin2021barf} and CamP~\cite{park2023camp} can improve photometric quality but mainly introduce small pose corrections and do not fully resolve these geometric discrepancies, suggesting that NeRF-based NVS-stereo needs additional pose optimization and RGB-D fusion. \textbf{(2) Artifacts:}
NeRF-generated views may contain floating or layered artifacts (Fig.~\ref{fig:challenges}B/C) that are not easily removed by simple filtering. Such artifacts introduce instability into depth supervision and downstream pose optimization. Applying conventional RGB-D fusion directly on unfiltered depth maps can therefore lead to under-converged or noisy reconstructions (Fig.~\ref{fig:challenges}D). \textbf{(3) Rendering–geometry trade-offs:} The quality of NeRF rendering directly impacts stereo depth. Depth-supervised iterative refinement has been explored~\cite{chen2024endoperfect}, but without joint pose optimization its gains can be limited in the presence of issues (1) and (2). Moreover, recent studies~\cite{wang2023digging,hu2024umednerf,rau2024depth} indicate that naively adding depth supervision may even degrade rendering quality, as depth and RGB losses can compete during optimization, highlighting the need for more carefully coupled objectives.

\section{Method}
\label{sec:method}

In our method, we adopt ZipNeRF as the backbone for NVS-stereo. ZipNeRF provides strong anti-aliasing and view-consistency properties, which ensure that the synthesized stereo pairs yield reliable initial depths for downstream optimization. To address the challenges in Sec.~\ref{sec:preliminary}, we propose a Multi-View Confidence-Guided Self-Supervised RGB-D Optimization (Mv-CG) framework with iterative depth–confidence voting and dynamic supervision updates. Our pose optimization framework is inherited from DROID-SLAM, but unlike the original DROID-SLAM, which treats depth as fixed, our method closes an EM-like (Expectation–Maximization) loop: after each bundle adjustment, we recompute cleaner, multi-view-consistent supervisory depth using updated poses and feed it back to the next BA round.



\textbf{Multi-View Depth Confidence Voting:}
Let there $I_{ref}$ be a reference frame with depth $D_{ref}$, intrinsics $K=[f_x,f_y,c_x,c_y]$, and extrinsics $(R_{ref},t_{ref})$. For $N$ neighboring frames $\{I_i,D_i,(R_i,t_i)\}_{i=1}^N$, we evaluate bidirectional consistency for each pixel $(u,v)$.

\textit{Forward projection} ($ref\!\rightarrow\!src_i$): A 3D point is unprojected from $(u,v)$ using $d_{ref}=D_{ref}(u,v)$:
\begin{equation}
\mathbf{p}^{ref}_{cam} =
\begin{bmatrix}
(u-c_x)\,d_{ref}/f_x\\
(v-c_y)\,d_{ref}/f_y\\
d_{ref}
\end{bmatrix},\quad
\mathbf{p}_{world} = R_{ref}^\top(\mathbf{p}^{ref}_{cam}-t_{ref}).
\end{equation}
It is then transformed to camera $i$ and projected:
\begin{equation}
\mathbf{p}^{src_i}_{cam}=R_i\mathbf{p}_{world}+t_i,\quad
\begin{bmatrix}u_{src}\\ v_{src}\end{bmatrix}
=\begin{bmatrix}f_x\,p_x^{src_i}/p_z^{src_i}+c_x\\[2pt]
                 f_y\,p_y^{src_i}/p_z^{src_i}+c_y\end{bmatrix}.
\end{equation}

Let $d_{src}^{\text{interp}}$ be bilinear interpolation of $D_i$ at $(u_{src},v_{src})$. Forward depth consistency holds if
\begin{equation}
\frac{\left|d_{src}^{\text{interp}} - p^{src_i}_{z}\right|}{p^{src_i}_{z}} < \tau_{\text{depth}}.
\label{eq:fwd_consistency}
\end{equation}

\textit{Backward projection} ($src_i\!\rightarrow\!ref$): The interpolated depth is back-projected and reprojected to the reference frame. The reprojection error $e_{\text{reproj}}$ and depth ratio must satisfy
\begin{equation}
e_{\text{reproj}}<\tau_{\text{reproj}}
\quad\text{and}\quad
\frac{\left|p^{ref,back}_{z}-d_{ref}\right|}{d_{ref}}<\tau_{\text{depth}}.
\label{eq:bwd_consistency}
\end{equation}
Bidirectional consistency is defined as
\begin{equation}
C_i(u,v)=\mathbf{1}\{\text{Eqs.~(\ref{eq:fwd_consistency}), (\ref{eq:bwd_consistency}) hold}\}.
\end{equation}

The total vote and confidence map are then
\begin{equation}
V(u,v)=\sum_{i=1}^N C_i(u,v),\qquad
\text{Confidence}(u,v)=\frac{V(u,v)}{N},
\end{equation}
and pixels supported by at least $m_{\text{vote}}$ views are retained:
{\small
\begin{equation}
M(u,v)=
\begin{cases}
1,& V(u,v)\ge m_{\text{vote}},\\
0,& \text{otherwise}.
\end{cases}
\qquad
D_{\text{filtered}}=M\odot D_{ref}.
\end{equation}
}

We initialize camera extrinsics with existing COLMAP results and perform an initial round of multi-view voting to obtain a coarse confidence map and filtered depth supervision, yielding a stable geometric prior that removes large artifacts before refinement. The voting loss is shown in Eq.~\eqref{eq:vote_loss}, minimizing \(L_{\text{vote}}\) maximizes the per-image ratio of pixels that pass the multi-view consistency threshold, thereby encouraging geometric consistency of the supervisory depth; here \(\Omega_r\) denotes the set of valid pixels in the reference image \(r\) (with \(|\Omega_r|\) its size), \(V_r(u,v)\) is the multi-view vote count at pixel \((u,v)\), \(m_{\text{vote}}\) is the pass threshold, \(\tau_m\) controls the softness of the decision, and \(s(x)=1/(1+e^{-x})\) is the logistic function.

\begin{equation}
L_{\text{vote}}
= 1 - \frac{1}{|\Omega_r|}
  \sum_{(u,v)\in\Omega_r}
  s\!\left(\frac{V_r(u,v)-m_{\text{vote}}}{\tau_m}\right),
\label{eq:vote_loss}
\end{equation}

\textbf{Confidence-Weighted RGB-D Optimization:}
Within each local DROID-SLAM window, we jointly optimize camera poses $\{\mathbf{T}_i\}$ and dense predicted depths $\{D^{pred}_i\}$ by minimizing
\begin{equation}
E = E_{\text{photo}} + E_{\text{depth}},
\end{equation}
where $E_{\text{photo}}$ is the standard photometric term and the RGB-D term is
{\small
\begin{equation}
E_{\text{depth}}
=\underbrace{\sum_{u,v} w(u,v)\,\rho\!\left(D^{pred}(u,v)-D^{obs}(u,v)\right)^2}_{\text{\scriptsize RGB-D term (optimizes poses)}}
+\lambda_{\text{vote}}\,L_{\text{vote}}.
\end{equation}
}

Here $D^{obs}$ is the current supervisory depth map (initialized from stereo using COLMAP poses), $\rho(\cdot)$ is a robust loss, and the weight
\begin{equation}
w(u,v)=\text{Confidence}(u,v)\cdot w_{\text{robust}}
\end{equation}
combines multi-view geometric confidence with robust iteratively reweighted least squares weighting. The total depth loss combines the confidence-weighted RGB-D supervision with the voting loss \(L_{\text{vote}}\), which encourages a higher proportion of pixels passing multi-view consistency.

\textbf{Iterative Supervision Refinement.}
After each local BA, we recompute multi-view consistency using the updated poses to obtain a refined stereo depth $\tilde{D}^{st}$ and confidence mask $M$. The supervisory depth is then updated via confidence-guided fusion:
{\scriptsize
\begin{equation}
D^{obs}_{t+1}(u,v)=
\begin{cases}
(1-\beta)\,D^{obs}_{t}(u,v) + \beta\,\tilde{D}^{st}_{t}(u,v), & \text{if } M_t(u,v)=1,\\[4pt]
D^{obs}_{t}(u,v), & \text{otherwise},
\end{cases}
\label{eq:fusion}
\end{equation}
}
where $\beta$ controls the update smoothness. To realize a coarse-to-fine schedule, we gradually anneal the consistency thresholds over iterations:
\begin{equation}
\tau^{(t+1)}_{\text{depth}} = \alpha\,\tau^{(t)}_{\text{depth}},\qquad
\tau^{(t+1)}_{\text{reproj}} = \alpha\,\tau^{(t)}_{\text{reproj}}.
\label{eq:anneal}
\end{equation}

As the poses converge, thresholds are gradually tightened, and the supervision update relies increasingly on the most reliable high-confidence pixels. 

\textbf{Global BA with NeRF Coupling:} Previous work shows that geometric RGB-D methods underperform COLMAP on NVS metrics~\cite{Azinovic_2022_CVPR}, while neural rendering-based SLAM systems incorporating NVS loss match or surpass COLMAP by directly reducing rotation errors~\cite{rosinol2023nerf,Zhu2022CVPR}. Concretely, on top of the global photo and depth optimazation objective, we attach a $1/8$-scale NeRF branch based on Nerfacto and adopt a hash-grid–aware coarse-to-fine schedule: Training first focuses on low-resolution hash levels with coarse proposal-guided sampling, then transitions to higher resolution and finer sampling to stabilize joint pose–radiance optimization.

\begingroup
\small
\setlength{\abovedisplayskip}{4pt}
\setlength{\belowdisplayskip}{4pt}
\setlength{\abovedisplayshortskip}{2pt}
\setlength{\belowdisplayshortskip}{2pt}
Let $\mathcal{S}_{1/8}$ denote $1/8$ spatial downsampling and $\mathcal{R}_{\theta}$ the NeRF renderer parameterized by $\theta$.
For each active frame $i$, we render $\hat I^{(1/8)}_i=\mathcal{R}_{\theta}(\mathbf{T}_i,K/8)$ and define the NeRF photometric loss (no confidence weighting) as
\begin{equation}
E_{\text{nerf}}
= \sum_{i\in\mathcal{A}} \;
  \sum_{\mathbf{x}\in\Omega^{(1/8)}}
  \rho\!\Big(
  \big\|\hat I^{(1/8)}_i(\mathbf{x})
  - \mathcal{S}_{1/8}(I_i)(\mathbf{x})\big\|_1
  \Big).
\end{equation}
The overall global objective becomes
\begin{equation}
E_{\text{global}}
= E_{\text{photo}} + E_{\text{depth}} + \lambda_{\text{nerf}}\,E_{\text{nerf}}.
\end{equation}
\endgroup

\begin{figure}[t]
    \centering
    \includegraphics[width=\linewidth]{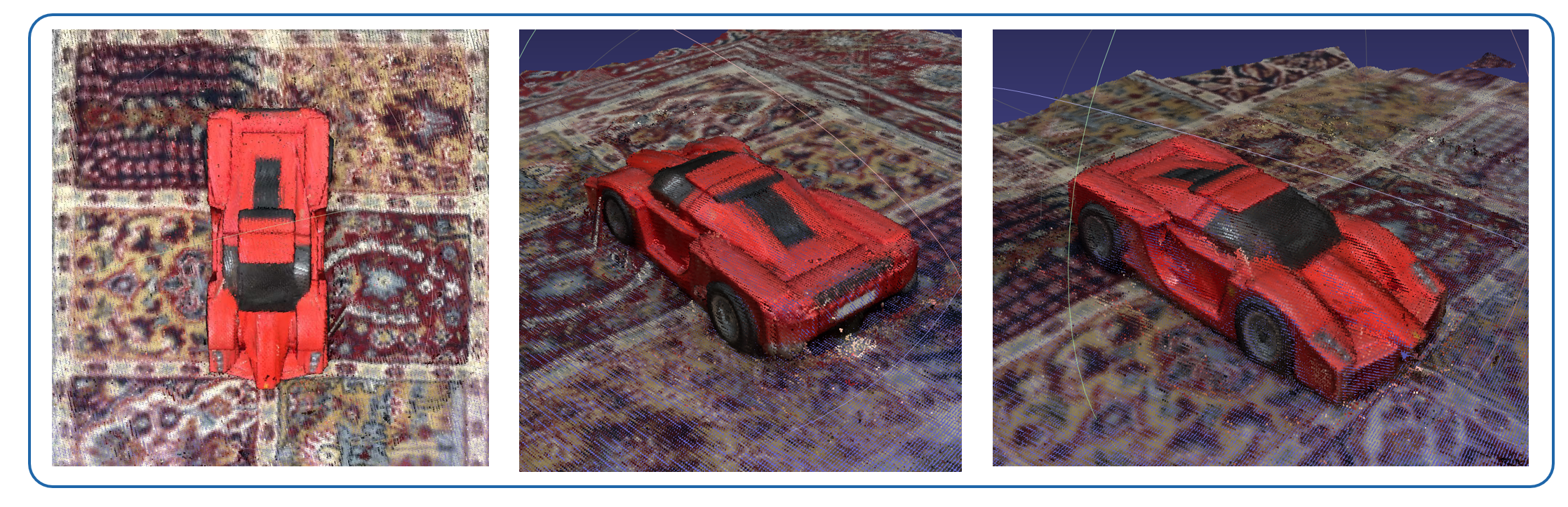}
    \caption{Pose optimization effectively eliminates the erroneous surface stacking shown in Fig.~\ref{fig:challenges}. By refining camera poses under our proposed mechanism, the projected depths from different views converge onto a coherent surface, yielding clean and well-aligned geometry across all viewpoints.}
    \label{fig:optimized}
\end{figure}

Within $\mathcal{R}_{\theta}$, the multi-resolution hash grid and proposal sampling naturally realize a coarse-to-fine behavior without per-pixel confidence reweighting, allowing long-range (loop) constraints, dense RGB-D geometry, and low-resolution NeRF photometrics to jointly regularize the same pose set $\{\mathbf{T}_i\}$ and prevent depth-induced pose degradation. With our proposed Mv-CG pose optimization mechanism, the stacking artifacts are largely eliminated. The produced well-aligned geometry illustrated in Fig.~\ref{fig:optimized}.

\textbf{Depth Fusion and NeRF Iteration:}
After the above optimization stages, we obtain refined camera poses together with a filtered stereo depth map. To fill holes and suppress flickering artifacts in stereo depth estimation~\cite{Bigdeli2014TemporallyCoherent}, we first perform TSDF fusion using the optimized poses and filtered depths. We then render depth maps from the fused TSDF mesh, align them with the filtered stereo depth, and use the TSDF depth to complete missing regions (see supplementary material for details). The resulting artifact-reduced, hole-filled depth combined with the refined poses is then used to address the rendering-quality limitation discussed in Sec.~\ref{sec:preliminary}. We adopt a depth-guided iterative strategy in the spirit of~\cite{chen2024endoperfect} to further improve ZipNeRF~\cite{barron2023zip}. Unlike~\cite{chen2024endoperfect}, we do not impose a direct depth loss, which may compete with the RGB objective. Instead, we guide ZipNeRF’s sampling using the optimized poses and completed depth, biasing samples along each ray toward the stereo-estimated surface geometry. For each pixel, we follow the standard NeRF convention and parameterize a camera ray as

\begin{equation}
\mathbf{r}(t)=\mathbf{o}+t\,\mathbf{d},\qquad t\in[t_n,t_f],
\label{eq:ray_param}
\end{equation}
where $\mathbf{o}$ is the camera origin, $\mathbf{d}$ is the unit viewing direction, and $t$ denotes the depth along the ray. The completed depth $\hat d$ simply specifies a preferred location along this ray for geometry-guided sampling.

{\scriptsize
\begin{equation}
p(t)=\frac{1}{Z}\exp\!\left(-\frac{(t-\hat d)^2}{2\sigma_d^2}\right),
\qquad
Z=\int_{t_n}^{t_f}\exp\!\left(-\frac{(s-\hat d)^2}{2\sigma_d^2}\right)ds,
\label{eq:gaussian_sampling}
\end{equation}
}
and draw sampling points using the inverse transform of the truncated Gaussian:
\begin{equation}
t_j = \hat d + \sigma_d\,\Phi^{-1}\!\big(\Phi_a + u_j(\Phi_b-\Phi_a)\big),
\qquad
u_j\sim\mathcal{U}(0,1),
\label{eq:inverse_transform_sampling}
\end{equation}
where $\Phi(\cdot)$ denotes the standard normal CDF and
$\Phi_a=\Phi\!\left(\frac{t_n-\hat d}{\sigma_d}\right)$,
$\Phi_b=\Phi\!\left(\frac{t_f-\hat d}{\sigma_d}\right)$.
This strategy concentrates sampling density around the estimated surface while keeping the ray sampling continuous and fully differentiable, enabling ZipNeRF to improve rendering quality and geometry without suffering from depth–RGB loss competition.

\textbf{Outputs:}
After the NeRF iteration stage, we obtain a refined ZipNeRF model and use it to synthesize stereo pairs with a substantially larger baseline than the initial
NVS-stereo setup. The increased stereo baseline yields larger disparity and therefore leads a more accurate depth estimates. Moreover, the refined ZipNeRF concentrates its sampling cones tightly around the true surface, leading to an explicit form of multi-view geometric consistency. As a result, the NVS-stereo depth no longer exhibits point-cloud stacking inconsistencies and can be directly fused into a mesh via TSDF fusion. An overview of the full pipeline is provided in Fig.~\ref{fig:framework}. Our system produces four main outputs: optimized camera poses, multi-view depth, the iteratively refined ZipNeRF, and the final reconstructed mesh. These outputs are highlighted with \textbf{green stars} in Fig.~\ref{fig:framework}, and their quantitative performance is evaluated in the subsequent experiments.

\section{Experiments}
\label{sec:exp}



We evaluate our method across four core tasks: camera pose estimation Sec.~\ref{Camera Pose Estimation}, multi-view depth estimation Sec.~\ref{Multiview Depth Estimation}, novel view synthesis~\ref{Novel View Synthesis} and surface reconstruction Sec.~\ref{Surface reconstruction}. Across all tasks, our method consistently achieves state-of-the-art performance compared with existing approaches. To further verify the effectiveness and robustness of our design, we conduct a series of ablation studies, including analyses of the impact of Mv-CG mechanism and NeRF coupling mechanism. Since earlier datasets were limited by video quality or ground truth accuracy, we adopt datasets collected within the last three years for our experiments. All ground-truth data are obtained either from high-accuracy sensors or synthetic sources with verified sub-millimeter accuracy. All the experiments are running on an H100 server; the setup details are shown in the \textbf{Supplementary Material}

\begin{figure*}
    \centering
    \includegraphics[width=0.9\linewidth]{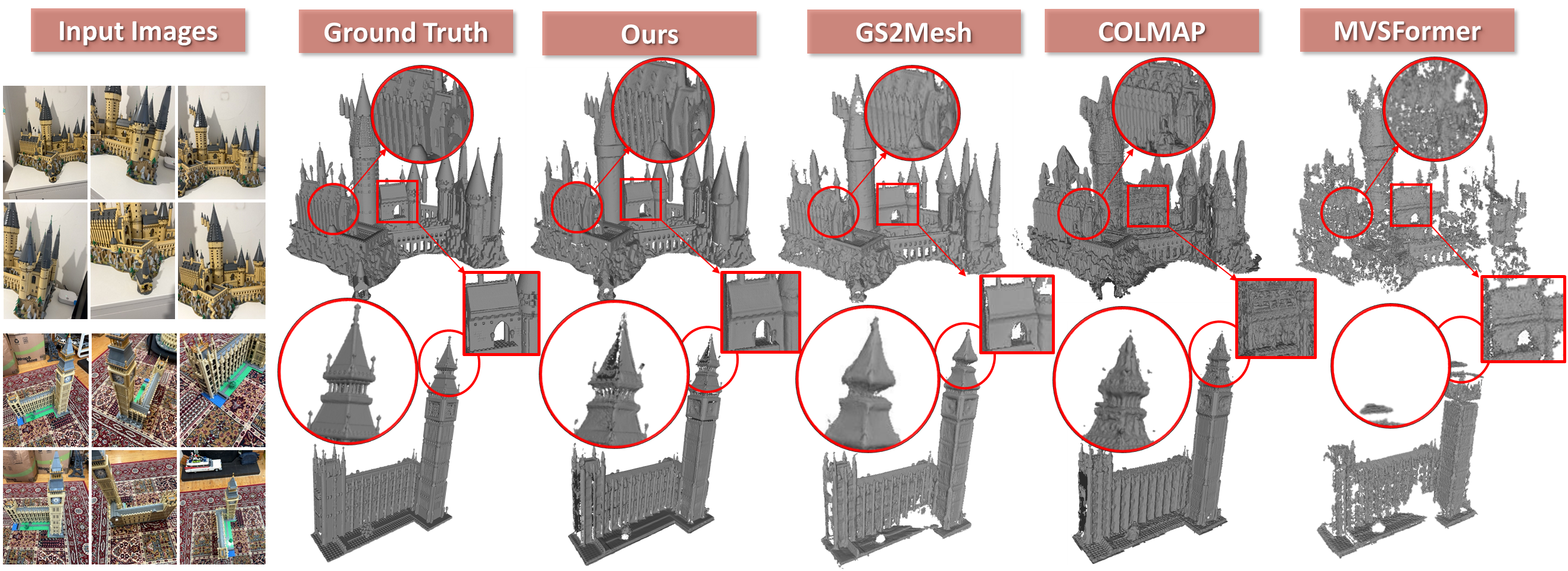}
    \caption{Mesh reconstruction comparison. With only multi-view RGB inputs, NeVStereo produces more accurate and geometrically faithful 3D reconstructions than competing methods.}
    \label{fig:mesh}
\end{figure*}
\subsection{Camera Pose Estimation} 
\label{Camera Pose Estimation}

\begin{figure}[!t]
    \centering
    \includegraphics[width=1\linewidth]{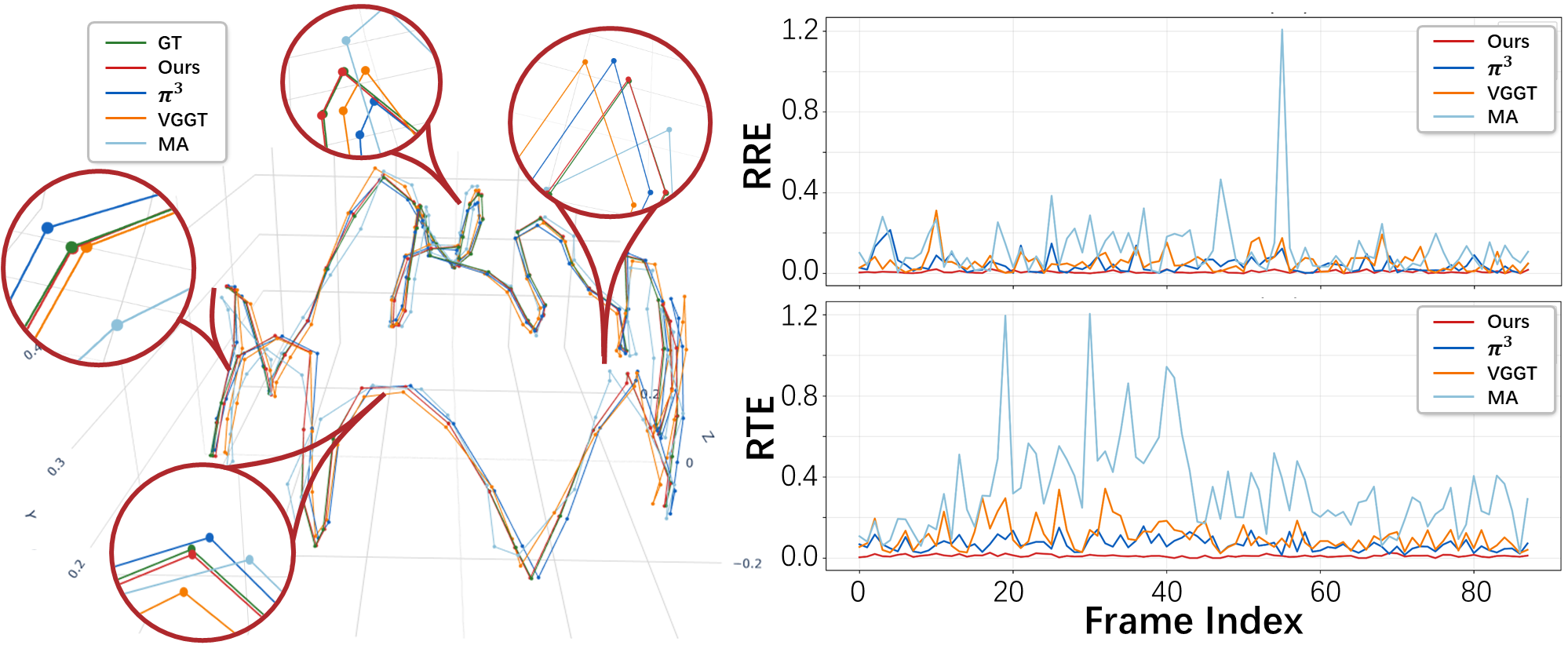}
    \caption{Trajectory and pose accuracy comparison. Left: estimated camera trajectory (red) nearly overlaps with the ground-truth trajectory (green). Right: per-frame RRE and RTE plots, where lower values indicate better pose accuracy.}
    \label{fig:pose}
\end{figure}

In this section, we feed all images into each feed-forward model in a single batch to obtain full trajectories for evaluation, as reported in Tab.~\ref{tab:hope_mobilebrick_comparison}. Under dense input, optimization-based methods such as VGGSfM~\cite{wang2024vggsfm} and COLMAP~\cite{wang2024vggsfm} remain highly competitive, particularly in rotation error. On small-scale, densely sampled scenes like Mobilebrick, feed-forward models show somewhat weaker performance relative to the optimization-based baselines. Our method outperforms both families across all metrics and surpasses the initial COLMAP poses, indicating that the proposed multi-view confidence–guided optimization and NeRF coupling can escape SfM local minima and further refine camera trajectories. We show the single frame's trajectory and pose accuracy comparison in Fig.~\ref{fig:pose}

\begin{table}[h]
\centering
\tiny
\setlength{\tabcolsep}{5pt}
\renewcommand{\arraystretch}{0.95}
\begin{tabular}{l|ccc|ccc}
\hline
\textbf{Method} &
\multicolumn{3}{c|}{\textbf{NVIDIA-HOPE(m)}} &
\multicolumn{3}{c}{\textbf{Mobilebrick(mm)}} \\
\cline{2-7}
 & \textbf{ATE}\,$\downarrow$ & \textbf{RTE}\,$\downarrow$ & \textbf{RRE°}\,$\downarrow$ 
 & \textbf{ATE}\,$\downarrow$ & \textbf{RTE}\,$\downarrow$ & \textbf{RRE°}\,$\downarrow$ \\
\hline
COLMAP~\cite{schoenberger2016sfm}  & 0.1461    & 0.0257          & \yellowbg 1.65    
        & \greenbg 1.852    & \yellowbg 0.967    & \yellowbg 0.0671 \\
VGGSFM~\cite{wang2024vggsfm}  & 0.1460    & \yellowbg 0.0246 & 1.67    
        & \yellowbg 2.064   & \greenbg 0.955    & \greenbg 0.0660 \\
VGGT~\cite{wang2025vggt}    & \redbg 0.1261    & \greenbg 0.0240    & 3.21    
        & 8.318    & 5.263    & 0.4389 \\
MapAnything~\cite{keetha2025mapanything}     
        & 0.1491    & 0.0271          & 2.02   
        & 21.11    & 14.748   & 1.7432 \\
$\pi^{3}$~\cite{wang2025pi}     & \yellowbg 0.1458    & 0.0352          & \greenbg 1.62    
        & 6.311    & 4.452    & 0.3621 \\
\textbf{Ours} 
        & \greenbg 0.1450    & \redbg 0.0215    & \redbg 1.61
        & \redbg 1.749       & \redbg 0.933     & \redbg 0.0649 \\
\hline
\end{tabular}
\caption{Average camera pose error across methods and datasets. {\redbg Best}, {\greenbg second-best}, and {\yellowbg third-best} results are highlighted.}
\label{tab:hope_mobilebrick_comparison}
\end{table}


\subsection{Multi-View Depth Estimation} 
\label{Multiview Depth Estimation}



\begin{table*}[htbp]
\centering
\scriptsize
\setlength{\tabcolsep}{5pt}
\renewcommand{\arraystretch}{0.95}
\begin{tabular}{l|cc|cc|cc|ccc}
\hline
\textbf{Method} &
\multicolumn{2}{c|}{\textbf{ScanNet++}} &
\multicolumn{2}{c|}{\textbf{Replica}} &
\multicolumn{2}{c|}{\textbf{NVIDIA-HOPE}} &
\multicolumn{3}{c}{\textbf{WildUAV}} \\
\cline{2-10}
 & \textbf{absrel}\,$\downarrow$ & $\boldsymbol{\delta{<}5\%}\,\uparrow$ 
 & \textbf{absrel}\,$\downarrow$ & $\boldsymbol{\delta{<}5\%}\,\uparrow$
 & \textbf{absrel}\,$\downarrow$ & $\boldsymbol{\delta{<}5\%}\,\uparrow$
 & \textbf{absrel}\,$\downarrow$ & $\boldsymbol{\delta{<}5\%}\,\uparrow$ & $\mathrm{Acc}_{\pm1\sigma}\,\downarrow$ (m) \\
\hline
DUSt3R~\cite{wang2024dust3r}       & 0.0359    & 86.66\%     & 0.0355    & 81.86\%     & 0.0631    & 65.28\%     & \greenbg 0.0269    & \greenbg 95.57\%     & \greenbg 0.311 \\
MonST3R~\cite{zhang2024monst3r}      & 0.0348    & 86.77\%     & 0.3653    & 13.29\%     & 0.1381    & 36.51\%     & 0.1251    & 42.59\%     & 3.306 \\
VGGT~\cite{wang2025vggt}         & \yellowbg 0.0312 & 84.47\% 
             & \greenbg 0.0096 & \greenbg 98.05\%
             & \yellowbg 0.0409 & 80.67\%
             &  0.0403 &  83.64\% &  0.703 \\
MapAnything~\cite{keetha2025mapanything}  & 0.0446 & \greenbg 87.37\%
             & 0.0274 & 94.60\%
             & 0.0483 & \yellowbg 81.75\%
             & 0.0444 & 82.69\% &  0.779 \\
$\pi^{3}$~\cite{wang2025pi}          & \greenbg 0.0256 & \yellowbg 86.79\%
             & \yellowbg 0.0102 & \yellowbg 97.67\%
             & \greenbg 0.0196 & \greenbg 93.62\%
             & \yellowbg 0.0338 & \yellowbg 91.58\% & \yellowbg 0.556 \\
\textbf{Ours}& \redbg 0.0163 & \redbg 94.87\%
             & \redbg 0.0054 & \redbg 99.25\%
             & \redbg 0.0134 & \redbg 97.82\%
             & \redbg 0.0039 & \redbg 99.51\% & \redbg 0.058 \\
\hline
\end{tabular}
\caption{Multi-view depth estimation comparison across ScanNet++, Replica, NVIDIA-HOPE, and WildUAV. 
Our method achieves the best absrel and $\delta{<}5\%$ accuracy on all datasets.}

\label{tab:mv_depth_scannetpp_replica_hope_wilduav}
\end{table*}

For the multi-view depth estimation task, we evaluate a range of state-of-the-art methods on four domains: synthetic (Replica~\cite{straub2019replica}), indoor (ScanNet++~\cite{yeshwanth2023scannet++}), tabletop (NVIDIA-HOPE~\cite{tyree2022hope}), and aerial (WildUAV~\cite{florea2021wilduav}), to assess generalization across scene scale, texture, and sensing conditions. Our experimental set includes the full Replica and NVIDIA-HOPE datasets and a ScanNet++ NVS-DSLR validation subset. Feed-forward methods are configured with a 3D reconstruction batch size of 25 (their empirically optimal setting), and we perform scale-invariant evaluation by applying only per-frame scale alignment. As reported in Tab.~\ref{tab:mv_depth_scannetpp_replica_hope_wilduav}, our method achieves the best absrel and $\delta{<}5\%$ scores across all benchmarks. Notably, on ScanNet++, we outperform DUSt3R~\cite{wang2024dust3r}, MonST3R~\cite{zhang2024monst3r}, VGGT~\cite{wang2025vggt}, and $\pi^{3}$~\cite{wang2025pi}, even though ScanNet++ is part of their training corpus. Point cloud visualizations in Fig.~\ref{fig:pointcloud} further indicate that our method tends to produce cleaner, more accurate geometry, whereas $\pi^{3}$~\cite{wang2025pi} and VGGT~\cite{wang2025vggt} often exhibit floating points and distortions, and MapAnything~\cite{keetha2025mapanything} shows more holes and warping.



\subsection{Novel View Synthesis}
\label{Novel View Synthesis}


\begin{figure}[t]
\centering
\includegraphics[width=\linewidth]{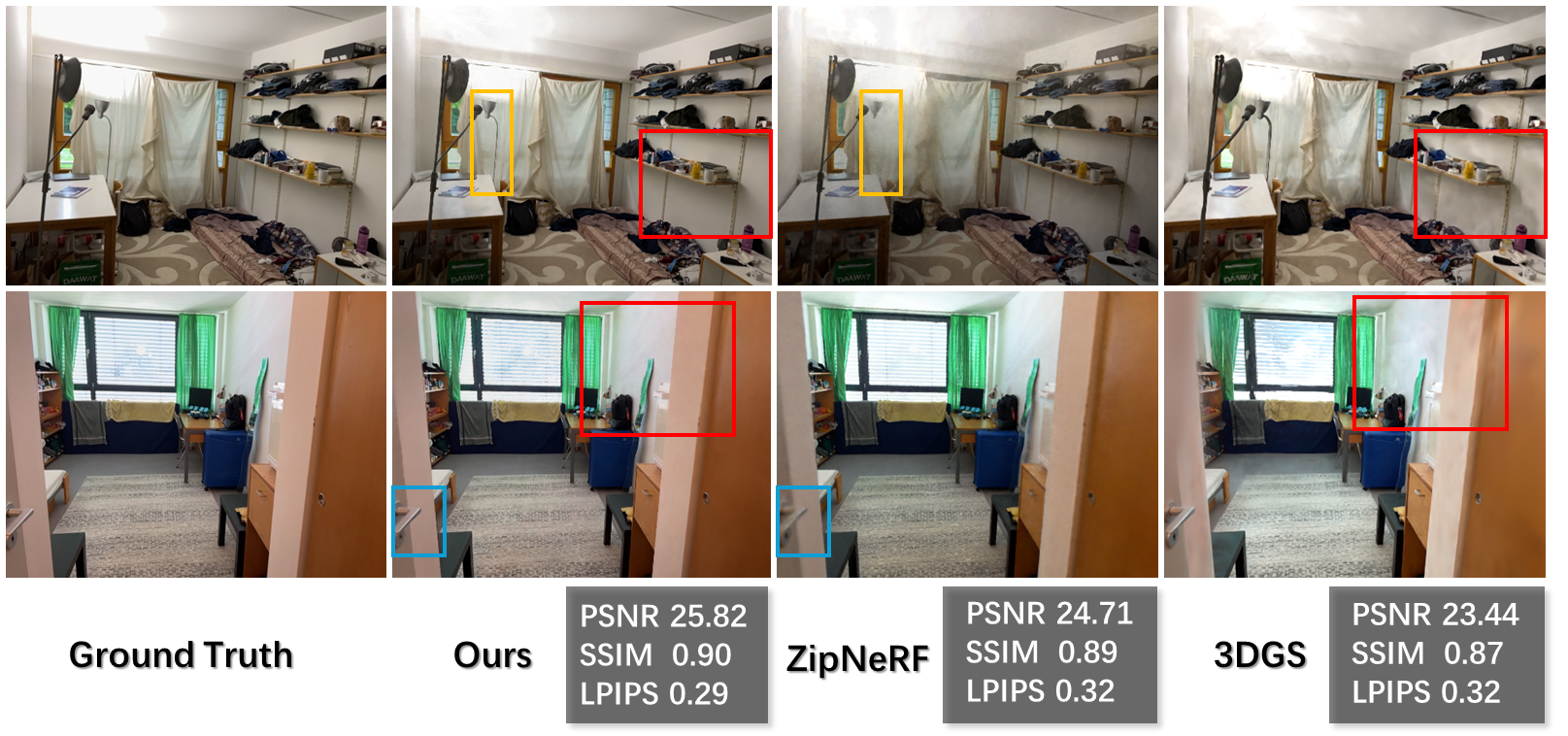}
\caption{Novel view synthesis comparison on ScanNet++. Our method produces sharper textures and more accurate geometry compared to baseline methods, including 3DGS and ZipNeRF.}
\label{fig:nvs_vis}
\end{figure}


We evaluate our method on the task of novel view synthesis using the same ScanNet++ validation subset. Our method in this section is the iteratively refined ZipNeRF, while baseline methods include vanilla 3DGS, Nerfacto, and vanilla ZipNeRF trained with COLMAP poses. Our method achieves the highest PSNR and SSIM and the lowest LPIPS, showing clear advantages in both photometric fidelity and perceptual quality compared with all baselines. Fig.~\ref{fig:nvs_vis} provides a visual comparison, demonstrating that our results are more photorealistic with noticeable improvements in color accuracy and fine details compared to the initial ZipNeRF.


\begin{table}[t]
\centering
\tiny
\setlength{\tabcolsep}{2.5pt}
\renewcommand{\arraystretch}{0.95}
\begin{tabular}{l|c|cc|cc|cc|c}
\hline
\textbf{Method} & \textbf{Input} &
\multicolumn{2}{c}{\textbf{Acc. (\%)}$\uparrow$  }& \multicolumn{2}{c}{\textbf{Rec. (\%)}$\uparrow$}&  \multicolumn{2}{c}{\textbf{F1}$\uparrow$}&
\textbf{Chamfer} \\
\cline{3-9}
 &  & $\sigma = 2.5$&$\sigma = 5$& $\sigma = 2.5$&$\sigma = 5$& $\sigma = 2.5$&$\sigma = 5$& \textbf{(mm)}$\downarrow$ \\
\hline
Neural-RGBD~\cite{Azinovic_2022_CVPR}  & I+D 
    & 20.61  &39.62 
& 10.66  &22.06 
& 13.67  &27.66 
& 22.78 \\
COLMAP~\cite{schoenberger2016sfm}  & I+ P& 74.89  &\yellowbg 93.79 
& 68.20  &84.53 
& \yellowbg 71.08  &\yellowbg 88.71 
& 5.26 \\
MVSFormer~\cite{cao2022mvsformer}  & I+ P& \greenbg 80.77  &\greenbg 96.33 
& 55.02  &71.32 
& 64.60  &81.14  
& 9.11 \\
NeuS~\cite{wang2021neus} & I+ P& \yellowbg 77.35  &93.33 
& \greenbg 70.85  &\yellowbg 86.11 
& \greenbg 73.74  &\greenbg 89.30 
& \greenbg 4.74 \\
SurfaceSplat~\cite{gao2025surfacesplat}  & I 
    & 69.61  &87.79 
& 68.89  &85.93 
& 69.14  &86.74 
& 9.90 \\
GS2Mesh~\cite{wolf2024gs2mesh} & I 
    & 68.77  &89.46 
& \yellowbg 69.27  &\redbg 87.37 
& 68.94  &88.28 
& \yellowbg 4.94 \\
\textbf{Ours} & I 
    & \textbf{\redbg 82.60}  &\textbf{\redbg 97.86} & \textbf{\redbg 71.49}  &\textbf{\greenbg 86.67} & \textbf{\redbg 76.64}  &\textbf{\redbg 91.93} & \textbf{\redbg 4.35} \\
\hline
\end{tabular}
\caption{
Multi-view reconstruction experiment ($\sigma{=}2.5$\,mm). Our method achieves the best overall accuracy, recall, F1, and Chamfer distance. Compared to neural-field-based approaches (e.g., NeuS, SurfaceSplat), MVS-based methods (e.g., COLMAP, MVSFormer) reconstruct accurately but can suffer from incomplete surface coverage. 
\textit{"I" denotes image input, "D" denotes requiring depth input, and "P" denotes requiring ground-truth camera poses.}
}
\label{tab:mv_reconstruction}
\end{table}

\subsection{Surface Reconstruction}
\label{Surface reconstruction}
We evaluate on Mobilebrick~\cite{li2023mobilebrick} by fusing our final depths and optimized poses. As summarized in Tab.~\ref{tab:mv_reconstruction}, our method achieves the best overall results. Qualitative results are shown in Fig.~\ref{fig:mesh}. As illustrated, our reconstructed meshes tend to be cleaner and more complete than those from traditional MVS pipelines (COLMAP with ground-truth poses and MVSFormer), with fewer artifacts and clearer structures. Compared with GS2Mesh, our method often recovers finer geometric details and exhibits more consistent surfaces.

\subsection{Abliation Study}


In this section, we investigate the effects of the confidence voting mechanism and NeRF coupling on both pose estimation and rendering quality. We perform a step-by-step analysis on the Mobilebrick dataset and additionally evaluate NeRF quality by holding out 5\% of camera views as a test set and training Nerfacto (without pose optimization) for 25K iterations. The results are summarized in Tab.~\ref{tab:mobilebrick_pose_nerf}. Directly optimizing poses with DROID-RGBD and unfiltered depth leads to degraded pose accuracy relative to COLMAP and also lower NeRF scores, indicating that naïvely applying RGB-D optimization can introduce both geometric drift and rendering instability. Introducing the Mv-CG mechanism improves all pose metrics and brings them close to COLMAP; however, the NeRF metrics slightly decrease, suggesting that more consistent local geometry does not necessarily translate into better global radiance-field reconstruction when pose and rendering remain decoupled. We also evaluate 3DGS-based NVS-stereo depth under the same Mv-CG framework and observe only marginal gains in both pose and NeRF metrics. A possible explanation is that 3DGS inherits COLMAP’s systematic depth biases, which Mv-CG cannot fully remove, limiting its corrective effect. In contrast, using NeRF-derived NVS-stereo depth within Mv-CG shows a different trend: NeRF’s implicit geometry is less tied to SfM biases, and under Mv-CG constraints it better approximates the true scene depth, improving both pose accuracy and NeRF scores. Finally, adding the NeRF coupling loss yields the best overall performance—pose errors are further reduced while PSNR/SSIM improve and LPIPS decreases—indicating that joint optimization of geometry, pose, and rendering provides a more stable and mutually reinforcing solution.



\begin{table}
\centering
\scriptsize
\setlength{\tabcolsep}{3.8pt}
\renewcommand{\arraystretch}{0.92}
\begin{tabular}{l|ccc|ccc}
\hline
\textbf{Method} &
\multicolumn{3}{c|}{Pose} &
\multicolumn{3}{c}{NeRF (Nerfacto)} \\
\cline{2-7}
 & ATE$\downarrow$ & RTE$\downarrow$ & RRE$\downarrow$
 & PSNR$\uparrow$ & SSIM$\uparrow$ & LPIPS$\downarrow$ \\
\hline
COLMAP  & \yellowbg 1.85 & \yellowbg 0.97 & \yellowbg 0.067 
        & \greenbg 24.51 & \redbg 0.85 & \yellowbg 0.203 \\

DROID (w/o Mv-CG) & 
        1.87 & 1.02 & 0.069 
        & 24.22 & 0.82 & 0.204 \\

DROID (w/ Mv-CG)  & 
        \greenbg 1.84 & \greenbg 0.95 & \greenbg 0.066 
        & 24.31 & \yellowbg 0.83 & \greenbg 0.201 \\

DROID + GS depth & 
        1.89 & 1.11 & 0.073 
        & 24.14 & 0.82 & 0.207 \\

\textbf{DROID-RGBD (Full)} & 
        \redbg 1.75 & \redbg 0.93 & \redbg 0.065 
        & \redbg 25.54 & \greenbg 0.84 & \redbg 0.195 \\
\hline
\end{tabular}
\caption{
Ablation study on the Mobilebrick. Mv-CG denotes Multi-View Confidence-Guided Self-Supervised Optimization. }
\label{tab:mobilebrick_pose_nerf}
\end{table}

\section{Conclusion \& Limitation}
\label{sec:conclusion}

We presented NeVStereo, a NeRF-based NVS-stereo framework that jointly outputs  high-accuracy pose, depth novel
view synthesis, and 3D reconstruction at the same time. By combining multi-view confidence–guided optimization, depth-guided iterative refinement, and NeRF-coupled bundle adjustment, our method attains strong performance in camera pose estimation, depth estimation, novel view synthesis, and surface reconstruction on multiple datasets. Our study suggests that NeRF-based NVS-stereo can provide more stable, pose-robust reconstructions than 3DGS-based counterparts, with promising zero-shot generalization even on datasets used to train competing methods. Nonetheless, our pipeline still depends on COLMAP initialization and reliable NVS quality, and performance may deteriorate under extremely sparse views. Future work includes exploring self-supervised initialization and improved robustness to view sparsity.
\newpage
{
    \small
    \bibliographystyle{ieeenat_fullname}
    \bibliography{main}
}


\end{document}